\documentclass[10pt,twocolumn,letterpaper]{article}

\usepackage{wacv}
\usepackage{times}
\usepackage{epsfig}
\usepackage{graphicx}
\usepackage{amsmath}
\usepackage{amssymb}
\usepackage{subcaption}
\usepackage{relsize}
\usepackage{caption}
\usepackage{multirow}
\usepackage{makecell}
\newcolumntype{L}[1]{>{\raggedright\let\newline\\\arraybackslash\hspace{0pt}}m{#1}}
\newcolumntype{C}[1]{>{\centering\let\newline\\\arraybackslash\hspace{0pt}}m{#1}}
\newcolumntype{R}[1]{>{\raggedleft\let\newline\\\arraybackslash\hspace{0pt}}m{#1}}
\newcommand{\J}{\mathcal{J}}
\newcommand{\F}{\mathcal{F}}

\DeclareMathOperator*{\argmax}{argmax} % thin space, limits underneath in displays

\wacvfinalcopy % *** Uncomment this line for the final submission

\def\wacvPaperID{****} % *** Enter the wacv Paper ID here

\ifwacvfinal\fi
\begin{document}

%%%%%%%%% TITLE
\title{RPM-Net: Robust Pixel-Level Matching Networks for \\ Self-Supervised Video Object Segmentation}

% \author{Youngeun Kim\\
% \and
\author{
Youngeun Kim\\
KAIST\\
\and
Seokeon Choi\\
KAIST\\
\and
Hankyeol Lee\\
KAIST\\
\and
Taekyung Kim\\
KAIST\\
\and
Changick Kim\\
KAIST\\
\and
{\tt\small \{youngeunkim, seokeon, hankyeol, tkkim93, changick\}@kaist.ac.kr}
}
\maketitle
%\thispagestyle{empty}

%%%%%%%%% ABSTRACT
\begin{abstract}
%The most costly, but necessary feature for  fully-supervised video object segmentation.
  % 1
  In this paper, we introduce a self-supervised approach for video object segmentation  without human labeled data.
  %
%   In this paper, we introduce a self-supervised approach, a training scheme without any human labeled  data, for learning video object segmentation.
  %
  % 2
  Specifically, we present Robust Pixel-level Matching Networks (RPM-Net), a novel deep architecture that matches pixels between adjacent frames, using only color information from unlabeled videos for training. 
  %% 문장 구조
  % 3
  Technically, RPM-Net can be separated in two main modules.
  % 4
  The embedding module first projects input images into high dimensional embedding space.
  % 5
%   Then the embedded features are selectively sampled by a Matching Module that uses deformable convolutions.
  Then the matching module with deformable convolution layers matches pixels between reference and target frames based on the embedding features.
  Unlike previous methods using deformable convolution, our matching module adopts deformable convolution to focus on similar features in spatio-temporally neighboring pixels.  
  Our experiments show that the selective feature sampling improves the robustness to challenging problems in video object segmentation such as camera shake, fast motion, deformation, and occlusion.
  % 6
%   The whole architecture is end-to-end trainable and does not require additional processing at inference time.
%   The whole architecture is end-to-end trainable and does not require online training at inference time.
%   The whole architecture is end-to-end trainable without does not require additional processing at inference time.
  % 8
  Also, we carry out comprehensive experiments on three public datasets (i.e., DAVIS-2017, SegTrack-v2, and Youtube-Objects) and
  achieve state-of-the-art performance on self-supervised video object segmentation.
%   superior results are obtained when compared to the latest self-supervised method. 
  %
  Moreover, we significantly reduce the performance gap between self-supervised and fully-supervised video object segmentation (41.0\%  vs.\begin{tiny} \end{tiny}52.5\% on  DAVIS-2017 validation set).
  
\end{abstract}

%%%%%%%%% BODY TEXT

\section{Introduction}
Video object segmentation, segmenting a foreground object along an entire video sequence, is one of the challenging tasks in computer vision.
%
% Most of the previous works propose multi-step training (references), matching algorithm (references) and using optical flow (reference) to improve the performance with human labeled annotations.
%
Most of the previous work \cite{Maskrnn, Monet, Mga, Trident, Modulation, OSVOS, RGPM, Track, VPN} focus on increasing the performance with human labeled annotations.
% using multi-step training, optical flow, and future frames. 
% However, compared to other tasks using videos (e.g., video object tracking and video object detection), video object segmentation suffers from limited training annotations since generating segmentation masks requires pixel-wise classification. 
%
% However, compared to other tasks using videos (e.g., video object tracking and video object detection), video object segmentation suffers from generating annotations for training since it requires pixel-wise classification.
However, compared to other tasks using videos (e.g., video object tracking and video object detection), video object segmentation suffers from generating pixel-level annotations for every frame \cite{annotation1, annotation2}.
For example, the DAVIS dataset \cite{Davis2017}, which is the most widely used dataset in the video object segmentation, contains 4,219 manually pixel-wise annotated frames.
% For example, even though we use only 100 videos which contain 100 frames in each video for training, 10,000 image segmentation annotations are needed.
% For example, the DAVIS dataset \cite{Davis2017}, which is the most widely used dataset in the video object segmentation, contains 60 training video sequences with total 4,219 annotated frames.
%
% It means that one needs to manually annotate about 4,000 images to obtain only 60 videos for training.
%
% It means that the considerable amount of effort is required to manually annotate about 4,000 images to obtain only 60 videos for training.

%The aforementioned problem has been rarely studied [refer, Lucid??].
%%%%%%%%%%%%%%%%%%%%%

%***************************FIGURE 1**************************************
\begin{figure}[t]
     
     \centering
         \includegraphics[width=0.35\textwidth]{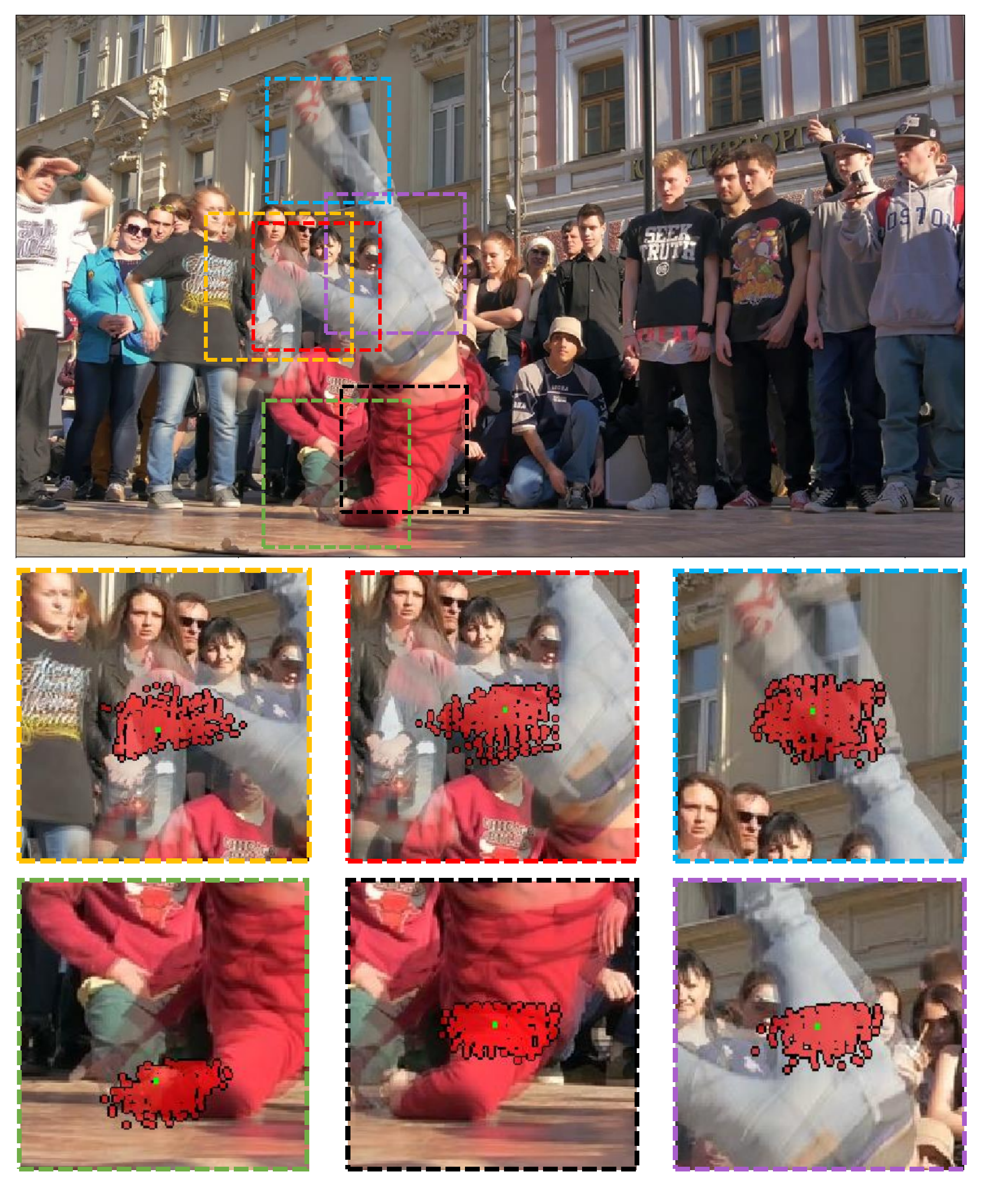}
     
     \caption{ Sampling locations (i.e., receptive fields of the matching module, marked in red) for target pixel (marked in green) matching are adaptively adjusted according to the change of frame.
     We overlay two consecutive frames for visualization.
     %
     %in challenging situations associated with video object segmentation: fast motion (top row), occlusion (middle row) and camera shaking (bottom row).
     %Lighter pixels indicate strong correlations with target pixel.
     The best view is in color and zoomed in. 
     }
     \label{Figure1}
\end{figure}
%****************************************************************************
The main motivation of this paper begins with relieving the amount of efforts to generate annotations for video object segmentation.
%%%%%%%%%%%%%%%%%%%%%
% Therefore, we tackle this annotation problem underlying video object segmentation.
%
% To this end, we introduce Robust Pixel-Level Matching Networks (RPM-Net) for self-supervised training in video object segmentation.
To this end, we introduce Robust Pixel-Level Matching Networks (RPM-Net) for self-supervised video object segmentation.
%
% Rather than using segmentation annotations, we use color information for network training since it is invariant to training process.
%
Rather than using segmentation annotations, RPM-Net leverages only color information for training.
% with the assumption that the color of adjacent frames is temporally stable.
%
The proposed network consists of two parts: an embedding module and a matching module.
The embedding module extracts high-dimensional features from input RGB images
and the matching module matches pixels between two frames according to embedding features.  
%
%%%%%%%%%%%%%%%%%%%%%

%%%%%%%%%%%%%%%%%%%%%
% In this paper, we tackle to this annotation problem underlying video object segmentation.
% %
% We introduce Robust Pixel-Level Matching Network (RPMN) for self-supervised training in video object segmentation.
% %
% Since there is no human supervision in training, the network is vulnerable to challenges in video object segmentation such as camera shaking, fast motion, deformation, and occlusion.
% %

%%%%%%%%%%%%%%%%%%%%%%%

\begin{figure}[t]
     \begin{subfigure}[b]{0.235\textwidth}
         \centering
         \includegraphics[width=\textwidth]{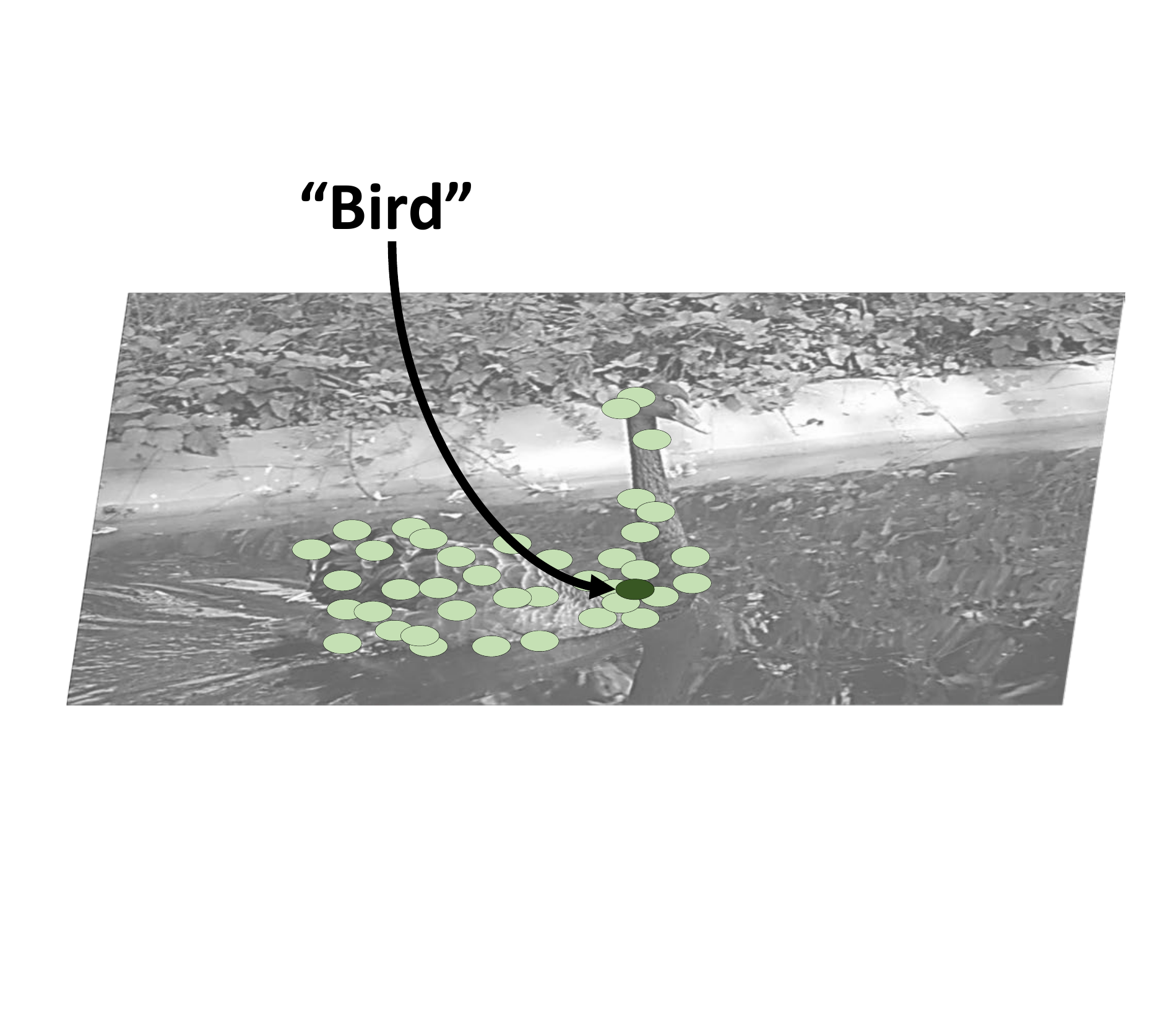}
         \caption{fully-supervised}
         \label{fig:analysis_graph_a}
     \end{subfigure}
     \begin{subfigure}[b]{0.235\textwidth}
         \centering
         \includegraphics[width=\textwidth]{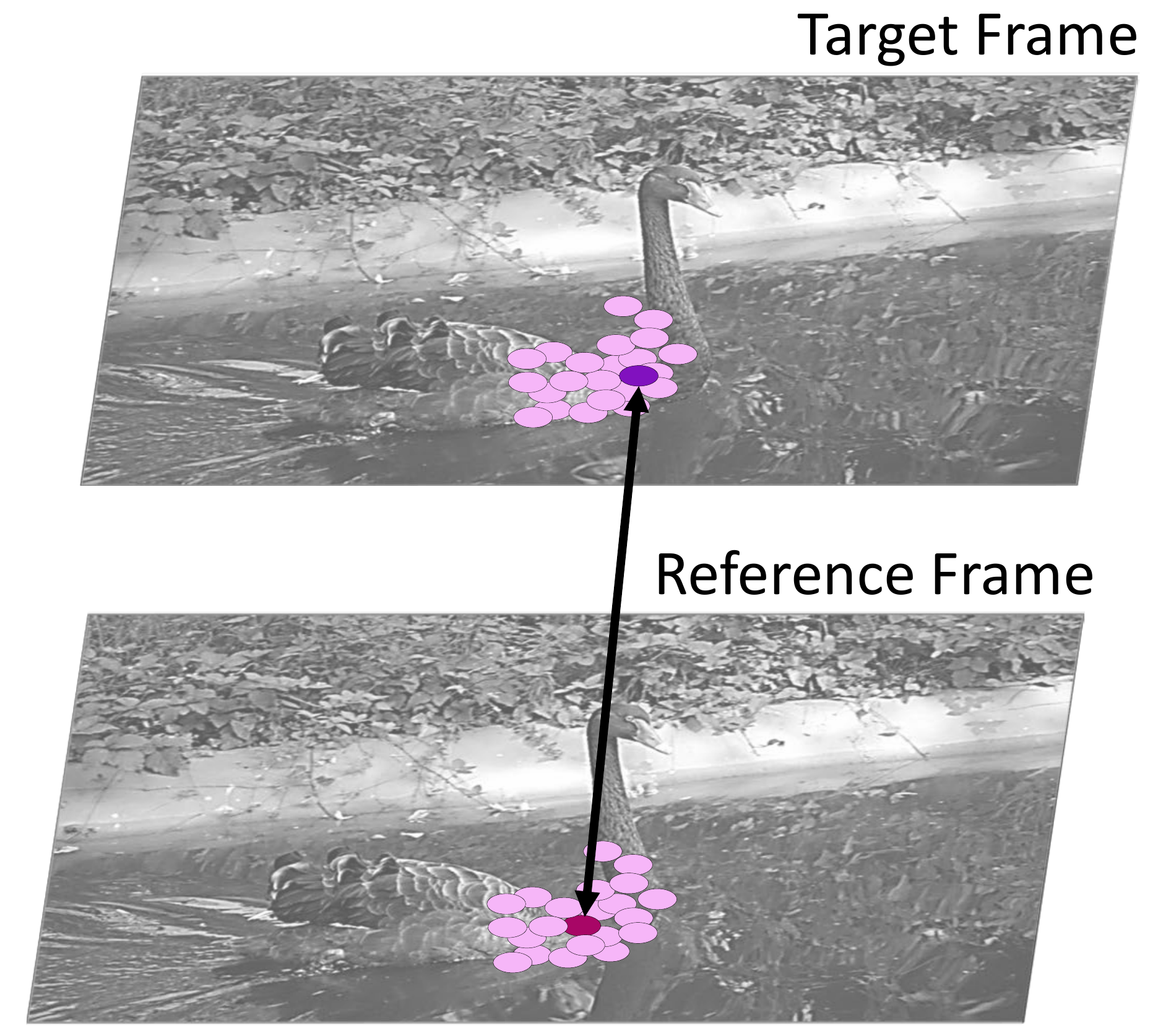}
         \caption{self-supervised}
         \label{fig:analysis_graph_b}
     \end{subfigure}
     \caption{ 
    %  The bus example shows that deformable convolution has different roles in supervised and self-supervised learning.
     Illustration of the role of deformable convolution in fully-supervised and our self-supervised learning.
     (a) 
     With human given class, the deformable convolution layers extract class geometric-invariant features for pixel-wise classification (we illustrate a segmentation example).  
     (b) 
     In our proposed method, the deformable convolution layers in the matching module focus on spatio-temporally neighboring pixels which have similar features.
     The best view is in color and zoomed in. 
     }
     \label{Figure2}
\end{figure}

%%%%%%%%%%%%%%%%%%%%%
However, since there is no human supervision for training, the network is vulnerable to challenges in video object segmentation such as camera shake, fast motion, deformation, and occlusion.
To address this problem, we use deformable convolution layers \cite{Deformable} in the matching module to focus on similar features in spatio-temporally neighboring locations (see Fig.\hspace{1ex}\ref{Figure1}), and show that this leads to robust pixel-level matching.
Moreover, we analyze how deformable convolution works in self-supervised learning with extensive experiments and visualizations.
In fully-supervised learning, deformable convolution discovers class \textit{geometric-invariant features} for semantic segmentation \cite{Deform_seg1, Deform_seg2}, object detection \cite{Deform_detect1, Deform_detect2, Deform_detect3}, and other computer vision tasks \cite{Deform_other1, Deform_other2, Deform_other3}.
%
% Unlike the conventional approaches, in our self-supervised scheme, deformable convolution in the matching module pays attention to similar features between the target pixel and its candidates in the reference frame (Figure  \ref{Figure2}).
Unlike the conventional approaches, in our self-supervised scheme, deformable convolution in the matching module pays attention to \textit{spatio-temporally similar features} (see Fig. \ref{Figure2}).
%``ploy"
% Unlike the conventional role, we use them to improve the robustness of pixel matching in self-supervised learning, which has no class-specific information to extract. 
%%%%%%%%%%%%%%%%%%%%%%%%%%%%%55
% Otherwise, 
% Since there is no class information in self-supervised learning, it is hard to match with class geometric-invariant features.
% %
% Instead, our attention matching network discovers  pixel-level invariant features between target pixel and its nearby matching pixels.  
% %
% Our experiments suggest that the matching module adjusts challenging situations with adaptive sampling locations and achieve higher performance compared to ones with fixed sampling locations. 
% %
%  Experiments and visualizations suggest that, although the network is trained without ground-truth labels, a mechanism for tracking automatically emerges.
%%%%%%%%%%%%%%%%%%%%%

%%%%%%%%%%%%%%%%%%%%%%%%%%%%%%%%
There are several advantages of RPM-Net:
% 1)
% Firstly, the proposed RPM-Net shows more accurate and reliable contour accuracy  ($\F$-score in the evaluation metric) compared to the latest self-supervised method \cite{TECV}. 
Firstly, the proposed RPM-Net can be trained in an end-to-end way with a single forward path, where all layers are differentiable.
%
%since our is the pixel-level matching in consecutive frames.
% 2)
%Even though RPM-Net achieves better performance,
% At the same time, we use about 3,000 videos for network training (refer), which is about 1\% of the number of videos used in previous work (refer) (300,000 videos (refer)).
% 3)
% Secondly, we can directly use RPM-Net at inference time without additional processes.
%
% In other words, there is no online training and pre-/post-processing during inference time.
% In other words, there is no online training and pre-/post-processing during inference time.
%
Secondly, although the model is only trained with unlabeled videos, RPM-Net well tracks the objects without online training at inference procedure,
which allows the network to operate with a high speed.
%and pre-/post-processing.
% 4)
% Moreover, RPM-Net segments target objects without accessing future frames, which allows RPM-Net can be used in a real-time application.
%
Moreover, RPM-Net can be used in real-world applications, since it segments target objects without accessing future frames.
To sum up, our contributions can be summarized as follows: 
1) 
We propose novel self-supervised video object segmentation framework, which is  annotation-free, end-to-end trainable, and no online training at inference process.
%
% Our method outperforms previous works, especially with respect to contour accuracy (32.7 vs 38.8  in DAVIS 2017 (refer) dataset).
%
2) In order to achieve that, we adopt deformable convolution \cite{Deformable} which is widely used in fully-supervised learning.
%
% Unlike the classical usage, we train them without labeling.
%
Our experiments and visualizations suggest that deformable convolution aggregates similar spatio-temporal features, so that the robust pixel matching is available.
%
% We give an intuition of how deformable convolution layers work in self-supervised learning.
%
To the best of our knowledge, this is the first time that deformable convolution is adopted in self-supervised learning and pixel-wise matching.
% 우리는 실험을 통해서 class가 없어 deformalbe이 잘 작동하는 것을 볼 수 있었다.
%
3) To provide a reference work on self-supervised video object segmentation,
% which is the desirable way to solve the annotation problem, 
and also to show the generality of RPM-Net,
we report the performance on three public video object segmentation datasets (DAVIS-2017  \cite{Davis2017}, SegTrack-v2 \cite{Segtrackv2}, and Youtube-Objects \cite{Youtube_Objects}).
%
% We outperforms the latest method on the DAVIS-2.
% Our RPM-Net shows state-of-the-art performance .
Our RPM-Net outperforms the latest self-supervised method.
% \cite{TECV}.
% , especially with respect to contour accuracy (32.7 vs. 38.8  on the DAVIS-2017 dataset).
%
% \hspace{0.01ex}
Most importantly, we significantly reduce the performance gap between self-supervised and fully-supervised video object segmentation.
%%%%%%%%%%%%%%%%%%%%%%%%%%%%%%%%%%%%%%%%5

%%%%%%%%%%%%%%%%%%%%%%%%%%%%%%%%%%%%%%%%%%
The paper is organized as follows.
Section 2 presents the related work.
Section 3 describes the details of the proposed frameworks.
Section 4 shows the performance on video object segmentation datasets 
and also presents the analysis of RPM-Net.
%
% At the end of this paper, we conclude with our results and point out the future work.
Finally, Section 5 concludes the paper.

%***************************FIGURE 7**************************************
%****************************************************************************

% %***************************FIGURE 2**************************************
% \begin{figure}[t]
%      \centering
%          \includegraphics[width=0.44\textwidth]{figure2.pdf}
%      \caption{ 
%     %  The bus example shows that deformable convolution has different roles in supervised and self-supervised learning.
%      The bus example shows the role of deformable convolution in fully-supervised and self-supervised learning.
%      %
%      In supervised learning  (left), with human given class, the deformable convolution layers extract class geometric-invariant features (marked in red). 
%      %
%      In our proposed method (right), without human supervision, the deformable convolution layers in matching module focus on spatio-temporal neighboring pixels which have similar features (marked in red).
%      %with a target pixel (green point).
%      %
%      The fully-supervised result directly obtained from \cite{Deformable}.
%      %
%      The best view is in color and zoomed in. 
%      }
%      \label{Figure2}
% \end{figure}
% %****************************************************************************
%***************************FIGURE 3**************************************
\begin{figure*}[t!]
     \centering
         \includegraphics[width=0.8\textwidth]{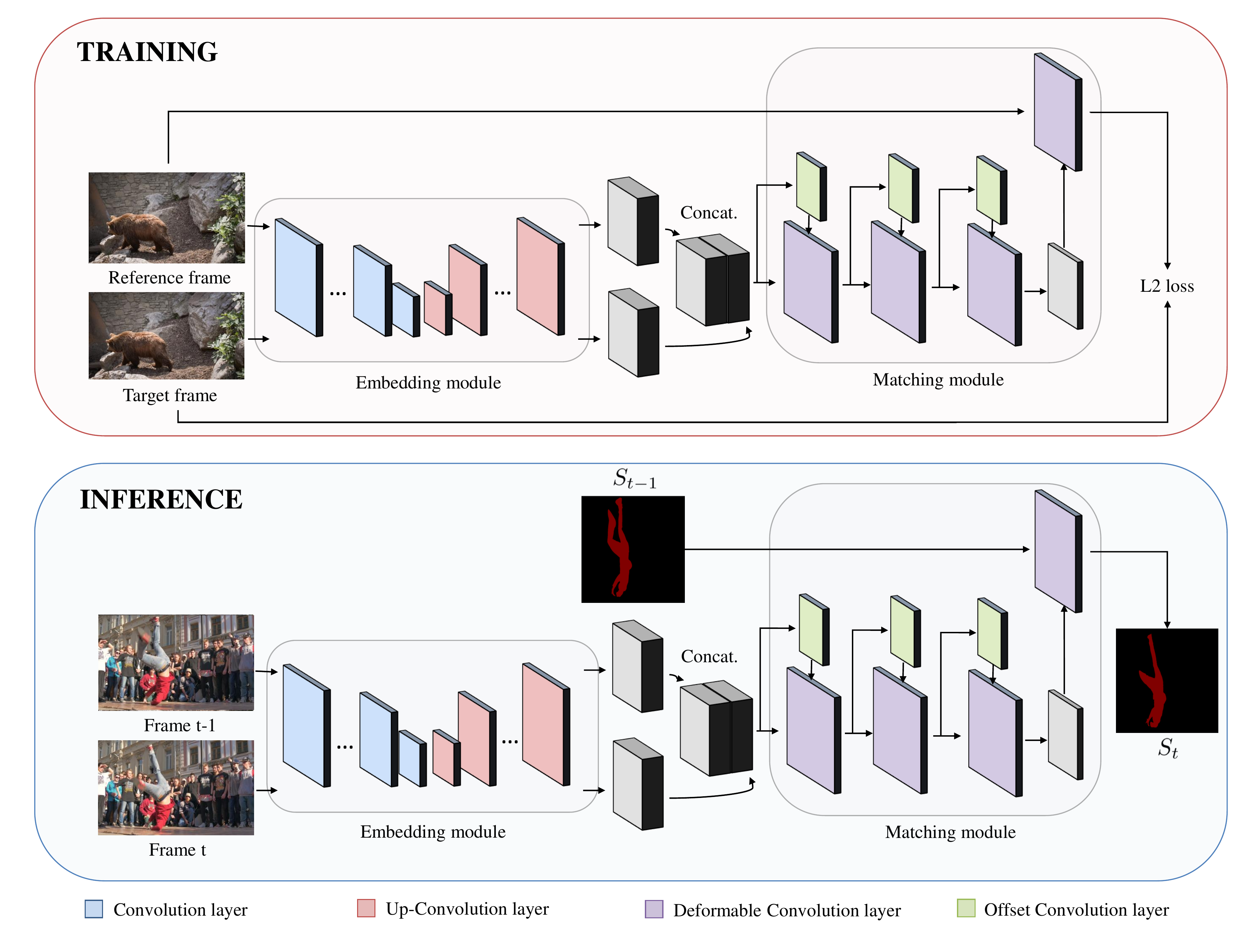}
     \caption{ The training and inference scheme of RPM-Net.
     The RPM-Net architecture consists of two parts: an \textit{embedding module} and a \textit{matching module}.
     The embedding module extracts features from input RGB images with up-convolutional layers to keep the resolution of embedding feature maps (128$\times $128). 
     The matching module is composed of deformable convolution layers and offset convolution layers.
     In our self-supervised training scheme, these layers are trained to focus on similar embedding features.
     %
    %  After training, the only thing RPM-Net needs to inference is to change input for matching module.
     %
    %  After training, RPM-Net infers the target segmentation mask $S_{t}$ using RGB images and segmentation mask $S_{t-1}$ of the frame $t-1$.
     After training, RPM-Net infers the target segmentation mask $S_{t}$ using RGB images and predicted segmentation mask $S_{t-1}$ of the frame $t-1$.  
     Note that a ground-truth segmentation mask is only given in the first frame.
     }
     \label{Figure3}
\end{figure*}
%****************************************************************************
%------------------------------------------------------------------------
\section{Related Work}

\subsection{Video Object Segmentation}

There is a large literature on video object segmentation since it is one of the most important tasks in video analysis.
Recently, most research focus on unsupervised and semi-supervised video object segmentation.
%
% The both methods require segmentation masks for training. 
%
Unsupervised methods segment primary objects without any information of these objects.
Since there is no given target object information in inference time, they use optical flow \cite{Unsup1_optical, Unsup2_optical, Unsup4_optical} and saliency maps \cite{Unsup3_saliency} for obtaining information about the target object.
Note that even though they are referred to as ``unsupervised" methods, these methods require annotations for network training.
%%%%%%%%%%%%%%%%%%%%%``ploy"

%%%%%%%%%%%%%%%%%%%%%
Semi-supervised methods segment foreground objects where object mask is given in the first frame.
Recent semi-supervised approaches are based on deep neural networks \cite{VideoMatch,  OSVOS, Reinforcement,segflow,Masktrack, Maskrnn} using semantic information from the first frame
and achieve reliable performance on given datasets.    
Our method is included in semi-supervised video object segmentation.
% (i.e., segmenting with the first frame annotation).
%
However, unlike the above two methods, we do not use annotations for training.
%, which is valuable in terms of the reduction of cost for generating annotation masks. 
%commercial application.
%

\subsection{Self-Supervised Tracking}
Our method is comparable to recent work by Vondrick et al. \cite{TECV}, which proposes a self-supervised tracking method using gray-scale features from target and reference frames. 
They outperform the high-performance optical flow network \cite{Flownet2.0} on the video object segmentation dataset \cite{Davis2017}.
However, this method is vulnerable to object deformation and scale change during the video sequence since they use only the several frames in the beginning of the video for inferring the target frame mask.
Also, the boundary distortion problem can be occurred since the features of object boundary usually have a large variation during the video sequence. 
Our RPM-Net deals with this problem by pixel-wise matching between adjacent frames so that the more reliable segmentation results are achieved.
%

%%%%%%%%%%%%%%%%%%%%%%%%
\subsection{Deformable Convolution (Background)}
Deformable convolution \cite{Deformable} is designed to improve the model geometric transformation capability of CNNs \cite{Deform_seg1, Deform_detect1}.
It shifts standard grid sampling locations $R$ by adding 2D offsets.
For example, considering the $3\times3$ kernel, standard convolution operation at location $\mathbf{p}_0$ on the input feature map  $\mathbf{x}$ and the output feature map $\mathbf{y}$ becomes
%
%%%% Eq_1
\begin{equation}
    \label{eq_1}
        \mathbf{y}(\mathbf{p}_0) = \sum_{\mathbf{p}_n\in R} \mathbf{w}(\mathbf{p}_n)\cdot \mathbf{x}(\mathbf{p}_0+\mathbf{p}_n),    
\end{equation}
%%%%
%
where $R=\left \{ \left ( -1, -1 \right ), \left ( -1, 0 \right ),..., \left ( 0, 1 \right), \left ( 1, 1 \right )\right \}$.
We use the same notation as in \cite{Deformable} for consistency.
%%%%%%%%%%%%%%%%%%%%%%%%%%%%%%%%%%%%%%%%%%%%%%%%%%%%

%%%%%%%%%%%%%%%%%%%%%%%%%%%%%%%%%%%%%%%%%%%%%%%%%%%%
In deformable convolution, convolution operation is formulated as follows:
%
%%%% Eq_2
\begin{equation}
    \label{eq_2}
        \mathbf{y}(\mathbf{p}_0) = \sum_{\mathbf{p}_n\in R} \mathbf{w}(\mathbf{p}_n)\cdot \mathbf{x}(\mathbf{p}_0+\mathbf{p}_n+\Delta \mathbf{p}_n).    
\end{equation}
%%%%
%
Here, offsets for each sampling point are denoted as 
% $\left \{ \Delta \mathbf{p}_n|n = 1,...,N \right \}$, where $N = |R|$.
$\left \{ \Delta \mathbf{p}_n|n = 1,...,|R| \right \}$.
Offset $\Delta \mathbf{p_{n}}$ is assigned from the output of another convolution layer (i.e., offset convolution layer) which takes the feature map $\mathbf{x}$ as input.
Here, the shifted sampling location ($ \mathbf{p} = \mathbf{p}_0+\mathbf{p}_n+\Delta \mathbf{p}_n$)
is the fractional location.
Therefore, $\mathbf{x}(\mathbf{p})$ is calculated by bilinear interpolation as
%
%%%% Eq_3
\begin{equation}
    \label{eq_3}
        \mathbf{x}(\mathbf{p}) = \sum_{\mathbf{q}\in Z} K(\mathbf{q},\mathbf{p})\cdot \mathbf{x}(\mathbf{q}),
\end{equation}
%%%%
%
where $K(\cdot,\cdot )$ is the bilinear interpolation kernel,
$Z$ is the set of all integral spatial locations on the input feature map $\mathbf{x}$.
Moreover, the bilinear interpolation kernel consists of two one dimensional kernels as 
%
%%%% Eq_3
\begin{equation}
    \label{two_one_dimensional_kernel}
        K(\mathbf{q},\mathbf{p}) = k(q_x, p_x)\cdot k(q_y, p_y),
\end{equation}
%%%%
%
where $k(a,b) = max(0, 1-|a-b|)$.
Refer to \cite{Deformable} for more details.
%%%%%%%%%%%%%%%%%%%%%%%%%%%%%%55
In our method, deformable convolution is adopted in the matching module which helps to achieve robust pixel-level matching without training annotations.
%

%------------------------------------------------------------------------
\section{Proposed Method}
\subsection{Overview}
Our goal is to segment foreground objects along an entire video without annotations for training.
Rather than training with dense annotations, RPM-Net leverages color information from unlabeled videos.
Our proposed network architecture is illustrated in Fig. \ref{Figure3}.
%
% and the detailed layer-by-layer definition is shown in Table \ref{layer-by-layer}. 
% and please see Appendix A in supplementary material for detailed layer-by-layer definition.

% Also, our intention is not treating the network as a black box, but specifying how the network behaves in self-supervised learning.
%
Also, our intention is not just building the network without interpretation, but specifying how the network behaves in self-supervised learning.
Therefore, we divide the network into two parts: an embedding module and a matching module.
Moreover, we verify that these modules work according to their purpose by showing experiment results in Section 4.
%
% We construct each module for their objective and verify that these modules work according to their purpose by showing experiment results in Section 4.
%
% We denote detailed layer-by-layer definition of two modules in Table 1 and Table 2, respectively.
%
In the remainder of this section, we first describe each module, and then demonstrate training and inference procedures in detail.
\subsection{Embedding Module}

%-----------------------------------
%Objective
The purpose of the embedding module is extracting a deeper representation from RGB images for more reliable matching.
Our embedding module produces 64-dimensional embedding feature maps from two given RGB images.
The extracted embedding features are taken into the matching module with concatenation.
A reference frame $\textit{I}_{ref}$ and a target frame $\textit{I}_{tar}$ are given as the input for the embedding module.
The inputs are resized to a resolution of $512 \times 512$.
We use FCN-ResNet101 \cite{FCN} for our embedding module
and modify up-convolution layers to obtain $128 \times 128$ feature maps. 
%
% Our embedding module architecture is shown in Table \ref{Table1}.
%
The process of the embedding module can be formulated as follows:
%%%% Eq_4
\begin{equation}
    \label{eq_4}
        \widehat{g}= concat(E(\textit{I}_{ref}), E(\textit{I}_{tar})),
\end{equation}
%%%%
where $E\left ( \cdot  \right )$ represents the modified FCN-ResNet101 in our embedding module, and $\widehat{g}$  denotes the concatenated features obtained from two input frames $\textit{I}_{ref}$ and  $\textit{I}_{tar}$.
%---------------------------------------------------

%%%%%%%%%%%%%%%%%%%%%%%
%-----------------------Table1--------------------
\begin{table}[t]
\caption{The layer-by-layer definition of the matching module.
%
% Please see Appendix A in supplementary material for detail of the embedding module. 
%
Batch normalization and ReLU non-linearity are omitted for brevity.
%
% Note that the last 1$\times$1 deformable convolution layer (i.e., layer 43) has fixed weights.
$D_{1}$ takes $I_{M1}$ as input, and $D_{4}$ directly takes $I_{M2}$ as input.
Please see Appendix A in supplementary material for detailed layer-by-layer definition including the embedding module.
}
\begin{footnotesize}

\begin{tabular}{c|c|c}
\Xhline{3\arrayrulewidth}
           & Layer Description                & Output Tensor Dim.    \\
\Xhline{3\arrayrulewidth}
 $I_{M1}$          &   Concatenated features &   (64, H/4, W/4)   \\ \hline
 $I_{M2}$         &   \makecell{ 
                    Reference image (training) or\\
                    $t-1$ Segmentation mask (inference)}  &  
                    \makecell{  (3, H, W) or \\
                                (1, H, W)}                    \\ \hline
    $D_{1}$         &  \makecell{3$\times$3 deformable conv, 32, stride 1 \\
                    offset conv: 3$\times$3 conv, 18, stride 1} &   (32, H/4, W/4)          \\ \hline
    $D_{2}$         & \makecell{3$\times$3 deformable conv, 16, stride 1 \\
                     offset conv: 3$\times$3 conv, 18, stride 1 }   & (16, H/4, W/4)                      \\ \hline
    $D_{3}$         & \makecell{3$\times$3 deformable conv, 2, stride 1 \\
                    offset conv: 3$\times$3 conv, 18, stride 1 } & (2, H/4, W/4)                      \\ \hline
    $D_{4}$         & \makecell{1$\times$1 deformable conv, 3, stride 1\\
                    offset: feature maps from layer 108 \\
                   }& \makecell{(3, H/4, W/4) or \\(1, H/4, W/4)                      }\\ 
\Xhline{3\arrayrulewidth}
\end{tabular}
\end{footnotesize}

\label{layer-by-layer}   
\end{table}
%-----------------------------------------------------------------------

%2___________________________________________________
\begin{table*}[t]
\caption{Quantitative evaluation on the DAVIS-2017 validation set \cite{Davis2017}, SegTrack-v2 \cite{Segtrackv2}, and Youtube-Objects \cite{Youtube_Objects} dataset.
We group the methods depending on whether human supervision is used for training. 
Also, we report the both $\J$-score (mIOU) and $\F$-score (contour accuracy) on the DAVIS-2017 validation set.
Here, we denote the RPM-Net with the refinement stage as RPM-Net$_{R}$.
}
\centering
\begin{tabular}{l|c|ccc}
\Xhline{3\arrayrulewidth}
Method             & Annotations for Training* & DAVIS-2017      & SegTrack-v2  & Youtube-Objects \\ \Xhline{3\arrayrulewidth}
Optical Flow (Coarse-to-Fine)     & X & 13.0 / 15.1          & 31.9            & 54.2               \\
Optical Flow (FlowNet 2.0)     & X & 26.7 / 25.2      & 35.1            & 52.4               \\
Colorization \cite{TECV}     & X & 34.6 / 32.7          & 41.4            & 55.0               \\\hline
RPM-Net with Conv (Baseline)      & X & 31.7 / 33.5          & 41.6            & 55.2               \\
RPM-Net with Dilation (Baseline)      & X & 30.8 / 33.1          & 41.8            & 54.4               \\
RPM-Net    & X & {35.7 / 38.8}        & {45.2}         & {56.2}            \\
RPM-Net$_{R}$    & X & \textbf{41.0 / 42.2}         & \textbf{48.4}         & \textbf{57.4}            \\ 
\hline
Modulation \cite{Modulation} &O& 52.5 / 57.1  & -            & 69.0     \\
OSVOS \cite{OSVOS}& O & 56.6 / 63.9  & 65.4            & 78.3     \\
% MoNet (ref)     &o   & -                    & {72.4}  & {81.7}\\
\Xhline{3\arrayrulewidth}
\end{tabular}

\label{Table2}
\end{table*}
%___________________________________________________________________________

% %-----------------------------------------------------------------------

\subsection{Matching Module}
The matching module matches pixels between reference and target frames using deformable convolution layers.
%based on their embedding features.
%
% The detailed architecture of the matching module is shown in Table \ref{layer-by-layer}.
%
To simplify notations, we use $ D\left ( a,b \right )$ to represent a deformable convolution layer, where $a$ and $b$ denote the input feature map and offsets, respectively.
Also, we represent a convolution layer for deformable offsets as  $O\left ( \cdot \right )$.
Since the matching module consists of 4 cascaded deformable convolution layers, 
we can write the output feature map $\widehat{f}_{k+1}$ from the deformable convolution layer as follows:
%%%% Eq_5
\begin{equation}
    \label{eq_5}
        \widehat{f}_{k+1} = D_{k}(\widehat{f}_{k},   O_{k}(\widehat{f}_{k})),
\end{equation}
%%%%
%
where $\widehat{f}_{k}$ denotes the input feature map of the layer $k$ in the matching module.
% (i.e., $\widehat{f}_{0}$ is same as $\widehat{f}$ in eq. \ref{eq_4}).
Note that $\widehat{f}_{1}$ obtained from the concatenated embedding feature map $\widehat{g}$ in eq. \ref{eq_4}.

As shown in Table \ref{layer-by-layer},
%  the matching module consists of 4 cascaded layers.  
%
the first three layers are $3 \times 3$ deformable convolution layers, which sample useful features from the concatenated embeddings for pixel-level matching.
And the last layer is a $1 \times 1$ deformable convolution layer with fixed weights of value 1.
Also, we do not use the offset convolution layer $O_{4}$ for the last deformable convolution layer $D_{4}$, and the layer directly use the features $\widehat{f}_{4}$ for offsets.
Thus, the last layer matches each target pixel with a fractional location in the reference frame based on offsets $\widehat{f}_{4}$.

The input feature map for the last deformable convolution layer is varied according to the training and inference. 
In the training procedure, the layer directly takes as input the RGB image from the reference frame and predicts the target frame image:
%%%% Eq_6
\begin{equation}
    \label{training_1by1}
        \widehat{I}_{tar} = D_{4}(I_{ref},  \widehat{f}_{4}).
        % \mathbf{y} = D_{4}(\mathbf{x},  \widehat{f}_{3}).
\end{equation}
%%%%
Otherwise, in the inference procedure, the reference segmentation mask $S_{t-1}$ is used for computing target segmentation mask:  
%%%% Eq_6
\begin{equation}
    \label{inference_1by1}
        {S}_{t} = D_{4}(S_{t-1},  \widehat{f}_{4}).
        % \mathbf{y} = D_{4}(\mathbf{x},  \widehat{f}_{3}).
\end{equation}
%%%%

% % %-----------------------------------------------------------------------

\subsection{Training and Inference}
In training and inference procedures, we assume that the image intensity does not change significantly between adjacent frames.
Therefore, we can match pixels with similar features and predict that these pixels belong to the same object.

\textbf{Training.}
\hspace{0.5ex}
Since our goal is to train RPM-Net without annotations, we use unlabeled videos for training.
%
% Here, we use the color intensity with the assumption that the color of adjacent frames is temporally stable.
%
Therefore, we do not represent network outputs as binary values (e.g., background: 0, target: 1), which requires annotations for training.
% In other words, we do not represent network outputs as binary values (e.g., background: 0, target: 1), which requires annotations for training.
%which is  widely used in fully-supervised learning (refer).
%
% Instead, we use $L2$ loss to penalize the intensity difference between two frames.
Instead, we use the $L2$ loss to penalize the color difference between the target frame $I_{tar}$ and the predicted target frame $\widehat{I}_{tar}$ from eq. \ref{training_1by1}.
Our loss function $L$ has the form:
%------------------------------------------------------------------------
% %%%% Eq_7
% \begin{equation}
%     \label{eq_7}
%     L = \frac{1}{N}\sum_{\mathbf{p}}^{}\left \| I_{tar}(\mathbf{p}) -\widehat{I}_{tar}(\mathbf{p})   \right \|_{2}^{2},
%     \end{equation}
% %%%%
% where $\mathbf{p}$ is the pixel location and $N$ denotes the total number of pixels.
%%%% Eq_7
\begin{equation}
    \label{eq_7}
    L = \frac{1}{N}\sum_{x}^{}\sum_{y}^{}\left \| I_{tar}(x,y) -\widehat{I}_{tar}(x,y)   \right \|_{2}^{2},
    \end{equation}
%%%%
where $(x,y)$ is the pixel location and $N$ denotes the total number of pixels.
%%------------------------------------------------------------------------

\textbf{Inference.}
\hspace{0.5ex}
% In our self-supervised scheme, there are two important things that should be considered for successful pixel matching:
% Since the segmentation mask should contain the integer class value (e.g., background: 0, dog: 1, and cow: 2) at each pixel, there are two important things that should be considered for successful pixel matching:
%
% (1) Recent video segmentation datasets \cite{Davis2017, Segtrackv2} have multiple primary objects in the video sequence.
%
% (2) Bilinear interpolation kernel in the deformable convolution layer (eq. \ref{eq_3}) generates fractional value at the object boundary.
In video object segmentation, the segmentation mask contains the integer class value (e.g., background: 0, dog: 1, and cow: 2) at each pixel.
Therefore, a fractional pixel value obtained from bilinear interpolation kernel in the deformable convolution layer (eq. \ref{eq_3}) reduces the segmentation performance of RPM-Net.

%
% To address the above-mentioned issues,
To address this problem, 
we first separate the segmentation mask  $S_{t-1}$ into the binary maps $B_{c}$  of each class $c$ including background class.
Then the binary maps are given as a input to the matching module.
%`
After that, we select the object class with the highest value at each pixel position  $\mathbf{p}$.
%
% Therefore, the operation of the last deformable convolution layer in the matching module (eq.\ref{inference_1by1}) can be specified as follows:
%  As shown in Figure \ref{Figure3}, we simply change the input for the last layer in the matching module for predicting the target segmentation mask
 %
Therefore, in the last deformable convolution layer in the matching module, the target segmentation mask ${S}_{t}$ is calculated as follows:
%%%% Eq_7
\begin{equation}
    \label{eq_7}
       {S}_{t}(\mathbf{p})  = \argmax_{c \in \mathcal{C}} D_{4}(B_{c}(\mathbf{p}), \widehat{f}_{4}),
\end{equation}
%%%%
%
where C is the set of class labels in the video sequence,
and  ${S}_{t}(\mathbf{p})$ and $B_{c}(\mathbf{p})$ denote the pixel value of the segmentation mask ${S}_{t}$ and the class binary map $B_{c}$ at the location $\mathbf{p}$, respectively.
%
% Moreover, to remove mismatched pixels in segmentation results, we add a refinement stage using CRFs \cite{CRF}.
Moreover,  we add a refinement stage using dense CRFs \cite{CRF}.
%
% We denote the RPM-Net with the refinement stage as RPM-Net$_{R}$ in our performance comparision (Table \ref{Table2}).
%

%
Note that RPM-Net can be directly used at inference procedure without online training.
%additional process such as online training and  pre-/post-processing.
%
Also, our RPM-Net segments primary objects without accessing future frames (e.g., $I_{t+1}$), which allows RPM-Net can be applied in real-world applications.
%
%

%***************************FIGURE 4**************************************
\begin{figure*}[t]
     \centering
         \includegraphics[width=0.90\textwidth]{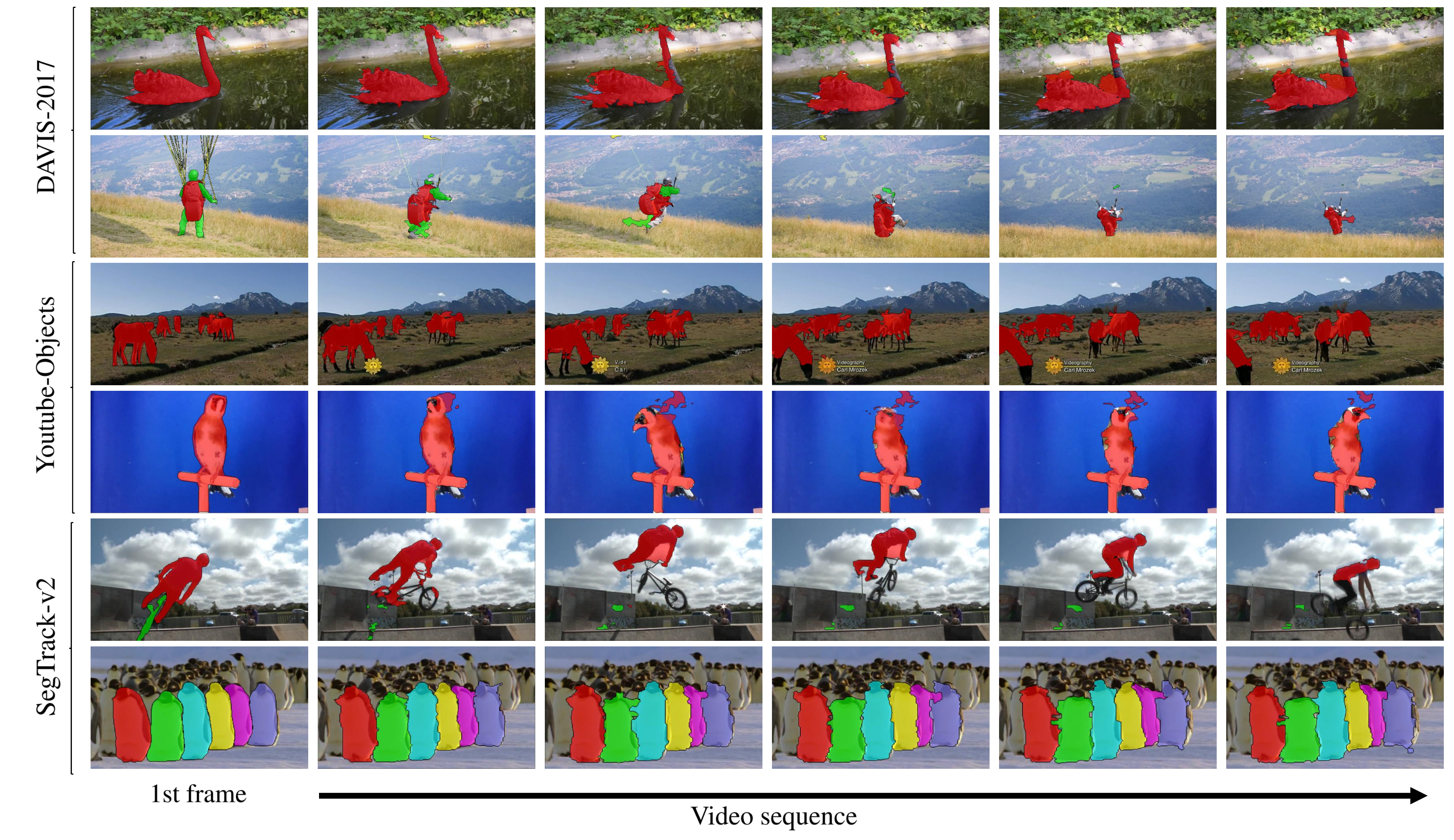}
     \caption{Qualitative results of RPM-Net (without the refinement stage) on three public video object segmentation datasets (i.e., DAVIS-2017, Youtube-Objects, and SegTrack-v2).
     }
     \label{qualitative results}
\end{figure*}
%****************************************************************************

%***************************FIGURE 5**************************************
\begin{figure*}[t]
     \centering
         \includegraphics[width=0.95\textwidth]{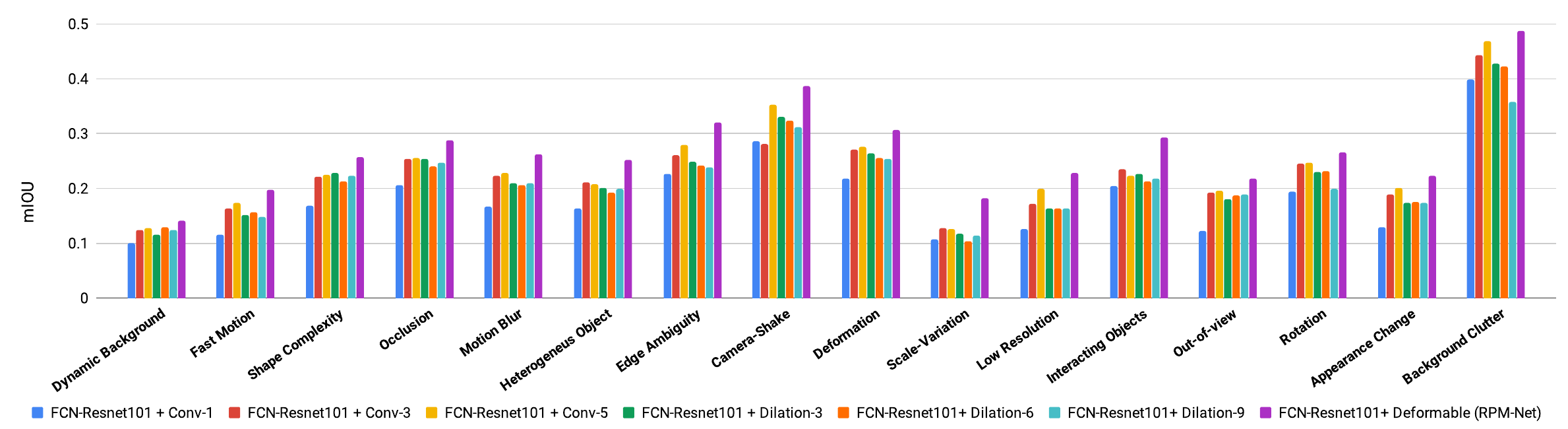}
     \caption{Visualization of attributes-based analysis on the DAVIS-2017 validation dataset. The matching module with deformable convolution achieves the best performance in all categories. In this experiments, we fix the embedding module to FCN-ResNet101.
     }
     \label{attribute_analysis}
\end{figure*}
%****************************************************************************

\section{Experimental Results}
In this section, we show the experimental results of RPM-Net.
%
% We first describe the datasets used for video object segmentation and evaluation metrics before we present quantitative and qualitative results, and the analysis of our proposed method.
%
We first describe the implementation details and the datasets used for video object segmentation
before we present evaluation metrics, quantitative and qualitative results, and the analysis of our proposed network.

\subsection{Implementation Details}
We train RPM-Net on the training set from Youtube-VOS \cite{Youtube_Large} and DAVIS-2016 \cite{DAVIS2016} (total 3,501 videos).
In the training procedure, we use frame $t-5$ as a reference frame and frame $t$ as a target frame.
%
% Moreover, to show the network can train without human supervision more clearly, we do not use pre-trained weights from other task.
%
Moreover, we do not leverage pre-trained weights since our ultimate goal is to train the network without any human supervision.
Therefore, the weights of all network layers are initialized with Gaussian distribution.
We train RPM-Net with 100 epochs on the training dataset and use the SGD optimizer with momentum 0.9, weight decay 0.0005, and initial learning rate 0.001. 
%
% Here, we adjust the ``ploy" learning rate policy \cite{Deeplab} for every epoch as $lr_{init}\cdot (1-\frac{epoch_{curr}}{epoch_{max}})^{0.9}$.
Here, we use ``ploy" learning rate policy \cite{Deeplab} for every epoch as $lr_{init}\cdot (1-\frac{epoch_{curr}}{epoch_{max}})^{0.9}$.

\subsection{Datasets}
To show the generality of our method, we report the performance on three public datasets (DAVIS-2017, SegTrack-v2, and Youtube-Objects).
%
% We use the same model and parameters for all experiments.
%

%
\textbf{DAVIS-2017.}
\hspace{0.5ex}
The DAVIS-2017 dataset \cite{Davis2017} is well-known video object segmentation dataset which includes multiple objects for each video sequence.
The dataset consists of 60 training videos and 30 validation videos.
%
%  To compare with the previous approach (refer), we also report the performance on DAVIS-2017 dataset.
Also, the dataset contains several videos from its previous version (i.e., DAVIS-2016 dataset).
However, it is more difficult to segment since multiple and similar objects appear in each video sequence. 

\textbf{SegTrack-v2.}
\hspace{0.5ex}
To validate the capability of RPM-Net, we further conduct further experiments on the SegTrack-v2 dataset \cite{Segtrackv2}.
The SegTrack-v2 dataset consists of 14 test video sequences with 24 objects.
The instance-level masks are available for multiple objects tracking along the video sequence.

\textbf{Youtube-Objects.}
\hspace{0.5ex}
The Youtube-Objects dataset \cite{Youtube_Objects} provides videos with 10 object categories (e.g., bird, boat, and train). 
Note that pixel-level segmentation masks are provided by \cite{Youtube-mask}.

\subsection{Evaluation Metrics}
We use the intersection over union metric  ($\J$) \cite{DAVIS2016} for evaluating region similarity.
The $\J$ score of DAVIS-2017, SegTrack-v2, and Youtube-Objects datasets are reported in Table \ref{Table2}.
Also, like the previous methods, we further compute contour accuracy ($\F$) \cite{DAVIS2016} for the DAVIS-2017 validation dataset.  

\subsection{Video Object Segmentation Results}
Table \ref{Table2} presents the performance of self-supervised and recent fully-supervised methods \cite{OSVOS, Modulation}  on the DAVIS-2017, SegTrack-v2, and Youtube-Objects datasets.
%
% Note that both methods segment foreground objects  where  object  mask  is  given  in the first  frame (i.e., semi-supervised video object segmentation). 
%
%
Firstly, we compared RPM-Net with the very recent self-supervised method \cite{TECV}.
As shown in Table \ref{Table2}, RPM-Net shows better performance in both region  similarity ($\J$) and contour accuracy ($\F$) on the DAVIS-2017 validation dataset.
%
% Especially, RPM-Net shows much better performance in contour accuracy.
% Especially, RPM-Net achieves about $18\%$ higher contour accuracy.
%
This is because \cite{TECV} uses only the several frames in the beginning of the video for inference, and also the features of object boundary usually have a large variation during the video sequence. 
Otherwise, RPM-Net deals with this problem by robust pixel-level matching between adjacent frames.
% so that the more reliable segmentation and boundary results are achieved.
%
Moreover, with the refinement step using CRFs, RPM-Net$_{R}$  further reduces the performance gap between self-supervised and fully-supervised video object segmentation.
Our qualitative results in Fig. \ref{qualitative results} show that although the model is only trained with unlabeled videos, RPM-Net achieves reasonable tracking results.

% Except for the DAVIS-2017 dataset, we additionally experiment on SegTrack-v2 and   
%
Obviously, there is a performance gap between fully-supervised and self-supervised way.
Nevertheless, through the self-supervised approach, the cost for human supervision is heavily decreased for training. 
Moreover, it is possible that further improvement in self-supervised learning might be achieved by more accurate pixel-level matching.

\subsection{Experimental Analysis}
In this section, we analyze the importance of each module with extensive experiments.
Before we present our experimental setup, a short note on terminology is required for clarity.
We use the ``sample locations" for referring the receptive fields of deformable convolutions, following the original paper \cite{Deformable}.
%
% Note that the sampling locations of deformable convolution are varied according to the feature map.
%
% Also, we use the theoretical receptive fields (refer) for experiments.  
%---------------------------------------------

%***************************FIGURE 7**************************************
\begin{figure}[t]
     \centering
         \includegraphics[width=0.36\textwidth]{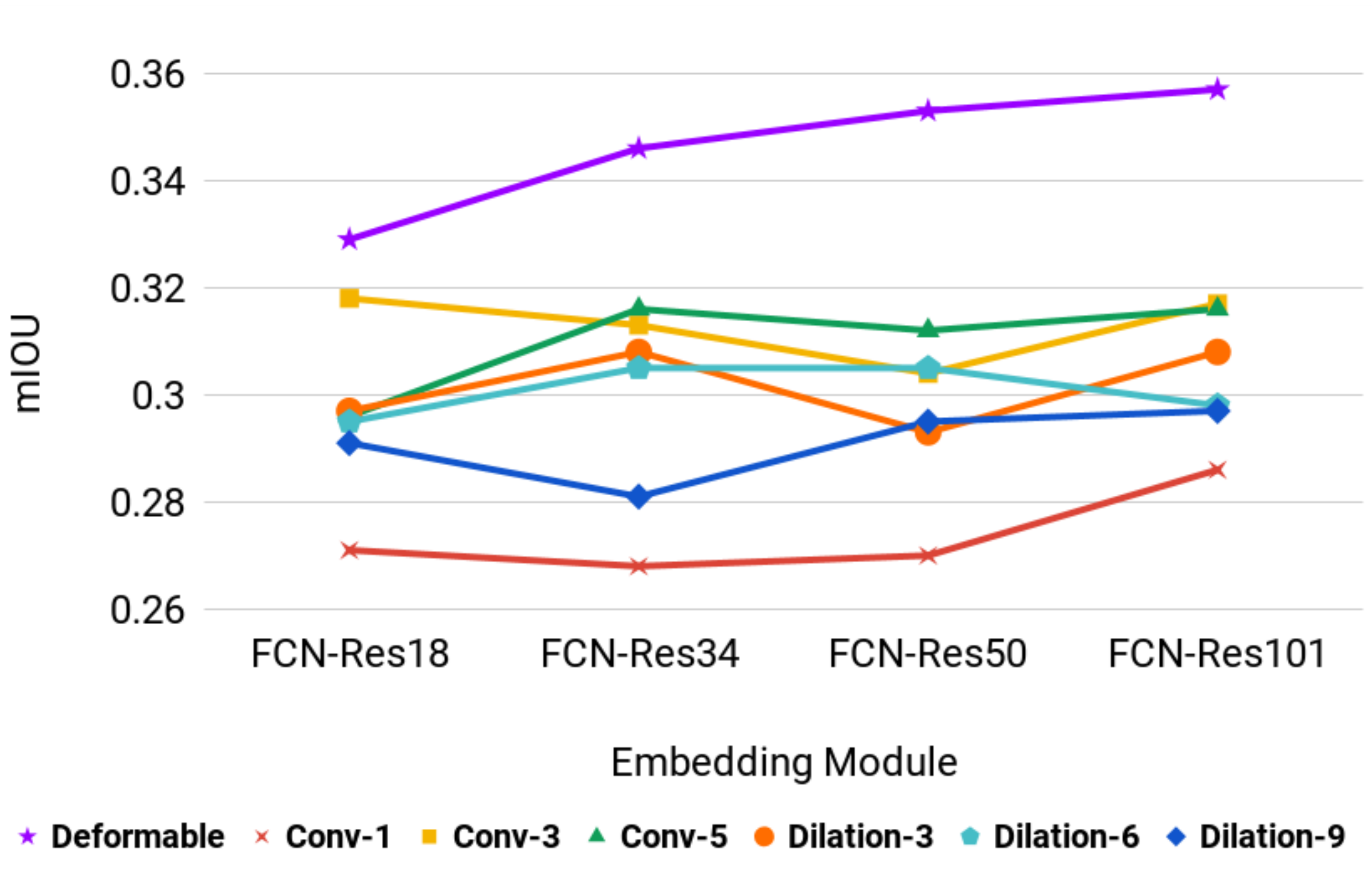}
     \caption{The importance of feature representation capability in the embedding module. We change the backbone network of the FCN model in the embedding module.
     }
     \label{mIOUvsEM}
\end{figure}
%****************************************************************************

%***************************FIGURE 7**************************************
\begin{figure}[t]
     \centering
         \includegraphics[width=0.38\textwidth]{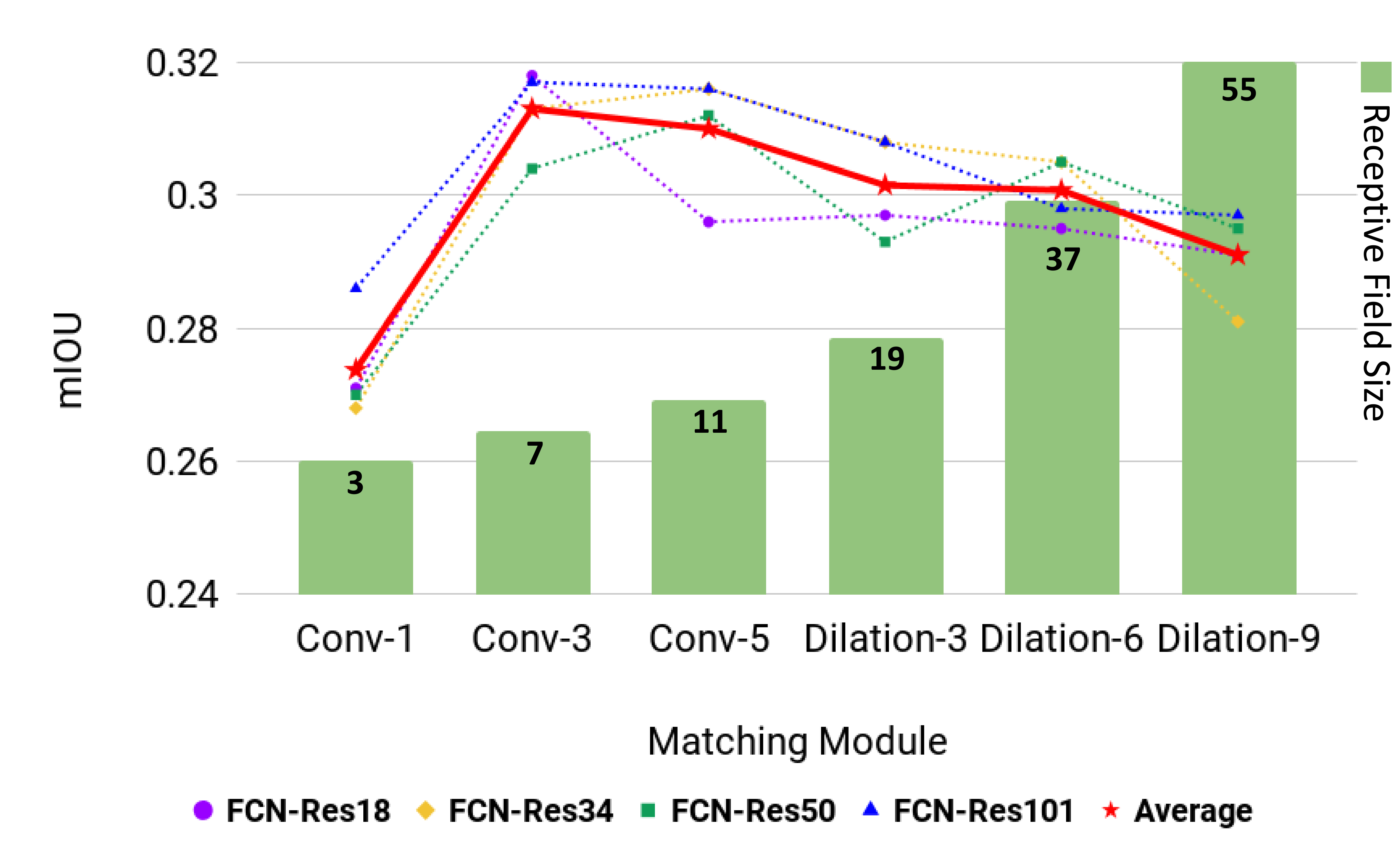}
     \caption{
     The relationship between performance and the receptive field size of the matching module.
     We overlay the theoretical size of receptive fields.
    %  The effect of receptive fields (or sampling locations) in the matching module. We overlay the theoretical size of receptive fields.
     }
     \label{mIOUvsRF}
\end{figure}
%****************************************************************************
\textbf{Experimental Setup.}
\hspace{0.5ex}
We experimented on the network with different combinations of two modules.
% as shown in Table \ref{Table4}.
%
In order to show the importance of feature representation capability in the embedding module,
%
% we used FCN-ResNet18, FCN-ResNet34, FCN-ResNet50, and FCN-ResNet101 for the embedding module.
we replaced the backbone network of FCN with ResNet-18, ResNet-34, and ResNet-50.
We also constructed matching modules using the standard convolution and dilated convolution \cite{Deeplab},
to present the effect of receptive fields in the matching module.
We varied the number of standard convolution layers ($L = 1, 3, 5$), and the dilation rate of dilated convolution layers ($D = 3, 6, 9$).
In the experiments, we fixed the number of dilated convolution layers to 3, which shows the best performance with dilated convolution.
Note that we maintain the last $1 \times 1$ deformable convolution layer for pixel-level matching.
All experiments were performed on the DAVIS-2017 validation set.
%
% In Figure \ref{attribute_analysis}, \ref{mIOUvsEM}, \ref{mIOUvsRF}, \ref{ablation_visual}, \ref{cosine_simalrity}  and Table \ref{time_comparison}, standard convolution with $k$ layer denotes as conv-$k$,
%
From now on, standard convolution with $k$ layers denotes as conv-$k$,
and also dilated convolution with dilation rate $r$ represents as dilation-$r$ for brevity.
Refer to Appendix B in supplementary material for quantitative performances of network configurations using in our experiments.

% -----------------------------------------------

\textbf{Embedding Module for Better Feature Representation.}
\hspace{0.5ex}
The embedding module generates high-dimensional features that help the robust matching.
As shown in Fig. \ref{mIOUvsEM}, the network performance is increased with deeper network architectures.
The results imply that rich feature information is essential for obtaining good performance in our self-supervised training scheme.
Moreover, the figure shows that deformable convolution achieves higher performance gain compared to others, which means that deformable convolution efficiently uses features from the embedding module.

\begin{figure}[t]
     \centering
         \includegraphics[width=0.47\textwidth]{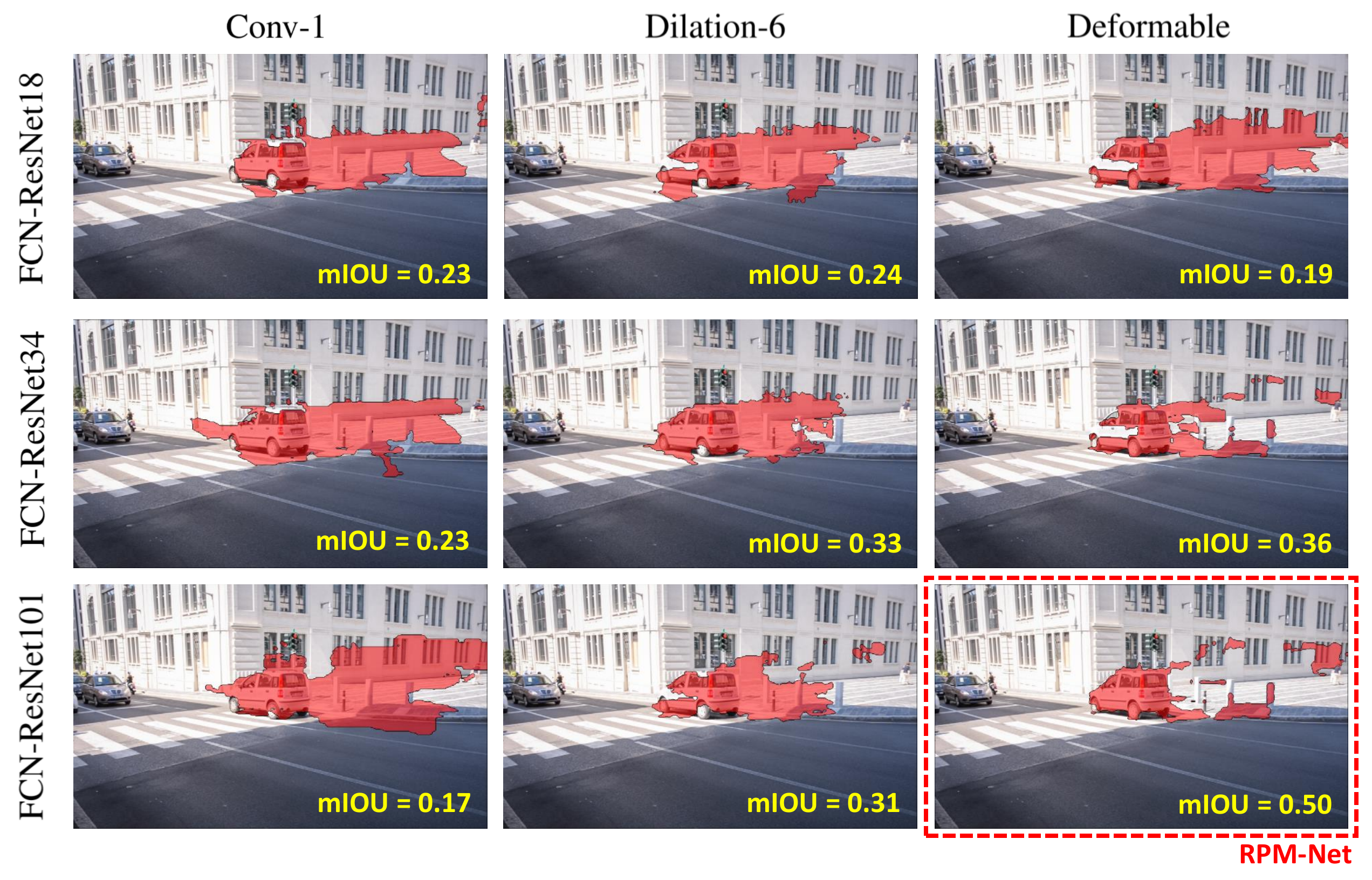}
     \caption{The effect of the embedding and matching modules.
     We visualize the last frame of the ``car-shadow" sequence in the DAVIS-2017 validation set.
     }
     \label{ablation_visual}
\end{figure}

\textbf{Matching Module for Robust Pixel-Level Matching.}
\hspace{4ex}
The performance of pixel-level matching is closely related with the receptive fields in the matching module.
Figure \ref{mIOUvsRF} shows the performance with varying the size of receptive fields.
In the figure, deformable convolution is excluded from the comparison, since it has adaptive receptive field size that can not be represented by fixed value.
In the figure, convolution with smallest receptive fields (i.e., conv-1) shows the lowest performance.
%
% This result demonstrates that small receptive fields rarely covers the reliable matching region.
This result demonstrates that the receptive fields size of 1-layer convolution is not enough to cover the matching region. 
% when the object undergoes   such as fast motion.
%since it rarely covers the objects with fast motion.  
%
However, the figure also shows that, except for conv-3, as the size of the receptive fields increases, the average performance is decreased.
We further discuss this observations in the following subsection (i.e., feature-level analysis of receptive fields).
% This results demonstrates that larger receptive fields is likely to contain redundant features for pixel matching.
%
In Fig. \ref{ablation_visual}, we compare the variants of our RPM-Net.

To show the effects of the matching module on challenging situations, we illustrate the performance of each video attribute \cite{DAVIS2016} in Fig. \ref{attribute_analysis}.
Deformable convolution achieves the best performance on the problems from the change of scene (i.e., fast motion, motion blur, camera shake, deformation, and scale-variation).
% when compared to others.
%
Also, it is beneficial for the problems from the objects configurations (i.e., heterogeneus object and interacting object).
For detailed description of the attributes, see \cite{DAVIS2016}.

%***************************Ablation Visual***********************************
%****************************************************************************

%***************************FIGURE 7**************************************
\begin{figure}[t]
     \centering
         \includegraphics[width=0.43\textwidth]{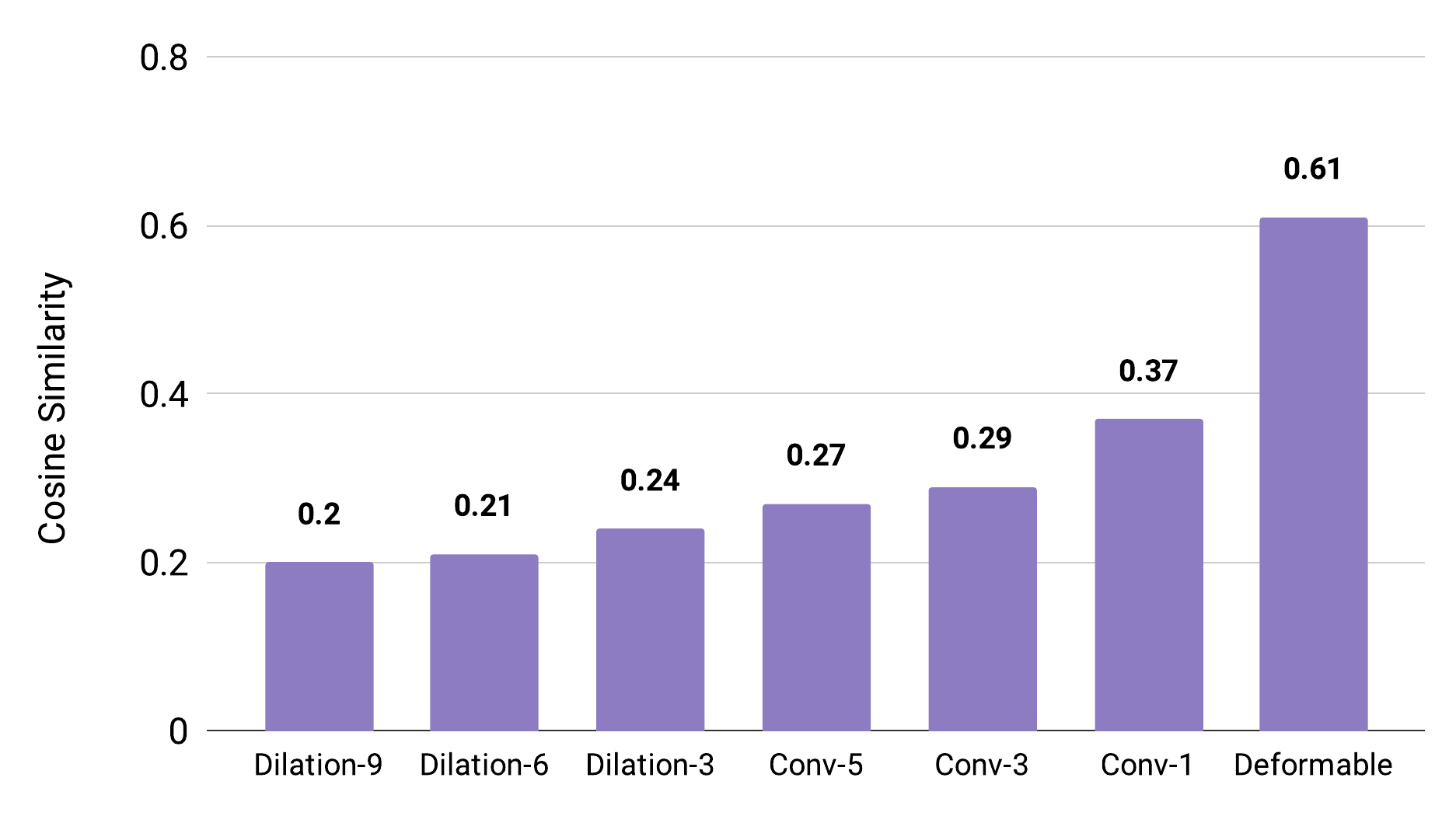}
     \caption{Feature similarities between embedding features in the receptive fields (or sampling locations).
     We use the DAVIS-2017 validation set for the experiments.
     }
     \label{cosine_simalrity}
\end{figure}
%****************************************************************************

\textbf{Feature-Level Analysis of Receptive Fields.}
\hspace{0.5ex}
%
% Moreover, we further analyze the embedding features that is contained in the receptive fields of the matching module.
Moreover, we further analyze the embedding features in the receptive fields of the matching module.
%
% To do this, in Figure \ref{cosine_simalrity},  we compute the average of cosine similarity between the feature at target location (sampled from the foreground objects) and features located in the receptive fields.
To do this, in Fig. \ref{cosine_simalrity},  we computed the average of cosine similarity between the feature vector at target pixel and other feature vectors located in the receptive fields.
%
% As we expected, convolutions with larger receptive fields show low similarity since it is likely to focus on redundant features.
The results demonstrate that larger receptive fields are likely to contain redundant features for pixel matching, which causes performance degradation as already shown in Fig. \ref{mIOUvsRF}.
Otherwise, deformable convolution shows significantly higher similarity value,
which means that the selective feature sampling is helpful to reduce the redundant features in the receptive fields.
% which means that the selective feature sampling is achieved by adopting deformable convolution.
%7392127137378

%***************************FIGURE 5**************************************
\begin{figure}[t]
     \centering
         \includegraphics[width=0.40\textwidth]{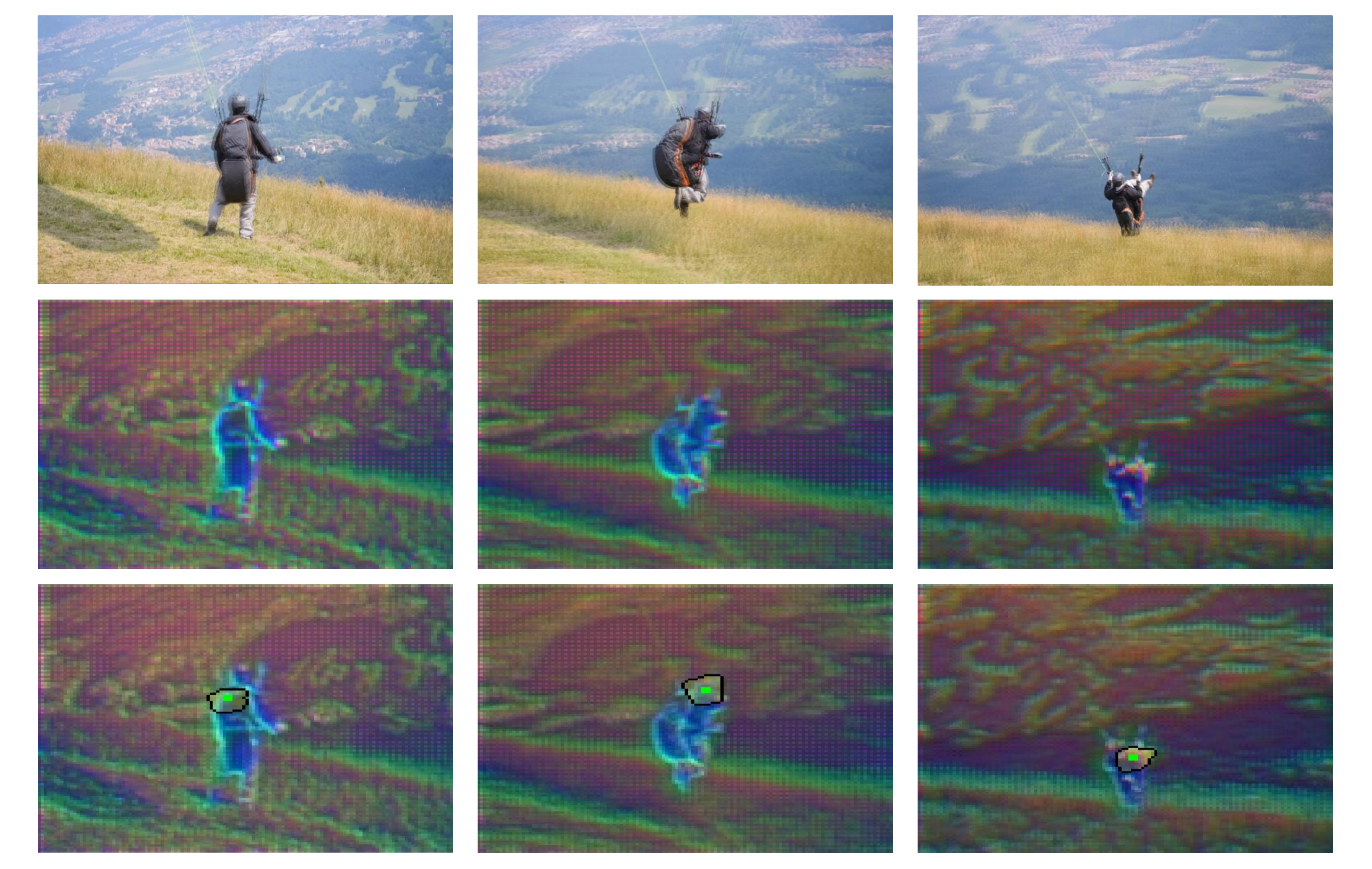}
     \caption{Visualization of the contributions of embedding and matching modules.
%      %
      For given frame $t-1$ and frame $t$ (two images are overlaid in the first row), 
      we show examples of embedding features (second row) and sampling locations (third row) for target pixel along the video sequence.
%      %
%      %
      The best view is in color and zoomed in. 
     }
     \label{module_visualize}
\end{figure}
%****************************************************************************

%----------------------------
\textbf{Visualizations from Two Modules.}
\hspace{0.5ex}
One may wonder whether our self-supervised training scheme is correctly implemented as we intended.
To alleviate this concern,  we present the results from two modules in Fig. \ref{module_visualize}.
For visualization, we project the 64-dimensional concatenated features (i.e., embeddings from frame $t-1$ and $t$) onto 3-dimensional space using PCA \cite{PCA}. 
From the results, we observe that the nearby pixels in the same object are represented in similar embedding features. 
Also, we overlay sample locations of the matching module on the projected embedding features.
It shows that the matching module enables selective feature sampling by focusing on similar embedding features.
%
%----------------------------

%***************************TABLE 5**************************************
\begin{table}[]
\caption{Computation time per frame on the DAVIS-2017 validation dataset.
% RPM$_{conv-3}$ denotes the 0 with 3 standard convolution layers in the matching module. 
}

\centering
\begin{tabular}{c|ccc}
                        \Xhline{3\arrayrulewidth}
        & RPM$_{conv-3}$ & RPM-Net & RPM-Net$_{R}$ \\ \Xhline{3\arrayrulewidth}
$\J$-score & 31.7       & 35.7    & 41.0       \\
$\F$-score & 33.5       & 38.8    & 42.2       \\ \hline
time       & 0.02$s$    & 0.09$s$   & 0.55$s$       \\
                        \Xhline{3\arrayrulewidth}
\end{tabular}
\label{time_comparison}
\end{table}
%****************************************************************************

%----------------------------
\textbf{Runtime of RPM-Net }
\hspace{0.5ex}
We evaluated the runtime of  RPM$_{conv-3}$, RPM-Net, and RPM-Net$_{R}$. 
%
% When we use standard and the dilated convolution for the matching module, the network runs at about 0.02s per frame.
%
As shown in Table \ref{time_comparison}, deformable convolution does not require large computation time. 
Also, we use 2 CRF iteration in RPM-Net$_{R}$ for obtaining reasonable speed.
% , the additional time consuming for CRFs is   
%
Although the network is trained without annotations, RPM-Net well tracks the objects apace.

%------------------------------------------------------------------------

\section{Conclusion}

% In this paper, we propose a novel self-supervised RPM-Net for video object segmentation.
In this paper, we have proposed a novel self-supervised RPM-Net for video object segmentation.
%
% We suggest the way to pixel-level tracking using only the color information of unlabeled videos.
Particularly, we adopted deformable convolution to improve the robustness to challenging situations in video.
Our experiments showed that RPM-Net provides reasonable tracking results without annotations.
Moreover, our analysis on two modules presented that
the proposed RPM-Net works well with deformable convolution in our self-supervised scheme.   
%
%how RPM-Net works with deformable convolution in our self-supervised scheme.
%
Our future work will be focused on the improvement of the robustness of the pixel matching in video object segmentation.

{\small
\bibliographystyle{ieee}
\bibliography{egbib}
}

\end{document}